\documentclass{article}

  \usepackage{wrapfig} 
\usepackage{subcaption}
\PassOptionsToPackage{numbers}{natbib}
\usepackage[final]{neurips_ai4d3_2025}

\usepackage[utf8]{inputenc} 
\usepackage{booktabs} 
\usepackage{amsmath}  

\usepackage[T1]{fontenc}    
\usepackage{hyperref}       
\usepackage{url}            
\usepackage{booktabs}       
\usepackage{amsfonts}       
\usepackage{nicefrac}       
\usepackage{microtype}      
\usepackage{xcolor}         
\usepackage{graphicx} 
\usepackage{algorithm}
\usepackage{algpseudocode}
\usepackage{amsmath}

\newcommand{\answerYes}[1][]{\textcolor{blue}{[Yes] #1}}

\newcommand{\answerNA}[1][]{\textcolor{gray}{[NA] #1}}
\newcommand{\answerTODO}[1][]{\textcolor{red}{\bf [TODO]}}

\title{Why Pool When You Can Flow? Active Learning with GFlowNets}

\author{%
  \begin{tabular}{cccc}
    \shortstack{Renfei Zhang$^{\diamondsuit}$\thanks{Equal contribution.} \\ \texttt{\small rza104@sfu.ca}} &
    \shortstack{Mohit Pandey$^{\spadesuit, \clubsuit}$\footnotemark[1] \\ \texttt{\small mkpandey@student.ubc.ca}} &
    \shortstack{Artem Cherkasov$^{\spadesuit, \heartsuit}$ \\ \texttt{\small artc@interchange.ubc.ca}} &
    \shortstack{Martin Ester$^{\diamondsuit}$ \\ \texttt{\small ester@sfu.ca}}
  \end{tabular}
  \\[0.95em]
  \normalfont
  $^{\diamondsuit}$\;School of Computer Science, Simon Fraser University, Burnaby, BC, Canada \\
  $^{\spadesuit}$\;Vancouver Prostate Centre, University of British Columbia, Vancouver, BC, Canada \\
  $^{\heartsuit}$\;Faculty of Medicine, University of British Columbia, Vancouver, BC, Canada \\
  $^{\clubsuit}$\;Diagen AI
}

\begin{document}

\maketitle

\begin{abstract}
 The scalability of pool-based active learning is limited by the computational cost of evaluating large unlabeled datasets, a challenge that is particularly acute in virtual screening for drug discovery.  While Active learning strategies such as Bayesian Active Learning by Disagreement (BALD) prioritize informative samples, it remains computationally intensive when scaled to libraries containing billions samples. In
this work, we introduce BALD-GFlowNet, a generative active learning framework that circumvents this issue. Our method leverages
Generative Flow Networks (GFlowNets) to directly sample objects in proportion to the BALD reward. By replacing traditional pool-based
acquisition with generative sampling, BALD-GFlowNet achieves scalability that is independent of the size of the unlabeled pool. In our virtual screening experiment, we show that BALD-GFlowNet achieves a performance comparable to that of standard BALD baseline while generating more structurally diverse molecules, offering a
promising direction for efficient and scalable molecular discovery.
\end{abstract}

\section{Introduction}
Active learning strategies are crucial for reducing the oracle labeling cost in large-scale machine learning tasks. However, traditional pool-based active learning faces significant challenges with scalability. Their reliance on scoring every instance in the unlabeled pool introduces a computational overhead that scales linearly with pool size, making them unsuitable for billion-scale datasets. 


These limitations are particularly acute in high-throughput virtual screening (HPVS) for drug discovery. Identifying small molecules with high binding affinity to a target protein is a critical step, but exhaustively evaluating massive molecular libraries via docking oracles is computationally prohibitive. For example, performing virtual screening on 1.3 billion ligands can take 28 days to complete, even with 8,000 GPUs \cite{gorgulla2020open, kim2024understanding}. 



To address the scalability limitation, we propose moving from a selective approach to a generative one. Instead of asking "Which samples from the current pool are most informative?", we ask "What does an informative sample look like?". To this end, we introduce \textbf{BALD-GFlowNet}, a generative active learning approach that integrates Generative Flow Networks (GFlowNets) with mutual information (MI)-based rewards \cite{houlsby2011bayesian, bengio2021flow, bengio2023gflownet, pandey2024gflownet}. Unlike traditional pool-based methods, BALD-GFlowNet directly generates informative samples guided by a reward function. This approach mitigates the scalability issue by decoupling the acquisition cost from the size of the unlabeled
pool. Our main contributions are:

\begin{itemize}
\item We introduce BALD-GFlowNet, a novel active learning framework that replaces pool-based acquisition with a generative policy trained to sample objects proportional to their BALD reward, making the acquisition cost independent of the unlabeled pool size.

\item We show on a synthetic grid task that BALD-GFlowNet identifies high-uncertainty regions more efficiently than exhaustive search, requiring fewer oracle calls to improve model performance. The learned GFlowNet policy demonstrates convergence towards high-uncertainty regions as training progresses, leading to better scalability without sacrificing the quality of acquired samples.

\item We apply BALD-GFlowNet to a large-scale virtual screening task for the Janus Kinase 2 (JAK2) protein \cite{gabler2013jak2}, showing it achieves performance comparable to the BALD baseline with superior efficiency. At a library size of 12 million, BALD-GFlowNet achieves a 12.5\% runtime reduction over the pool-based BALD baseline \citep{irwin2005zinc}. Here our method not only prioritizes informative samples but also ensures their chemical viability and structural diversity.
\end{itemize}

\begin{figure}
  \centering
\includegraphics[width=1.1\textwidth, trim=0 150 0 0, clip]{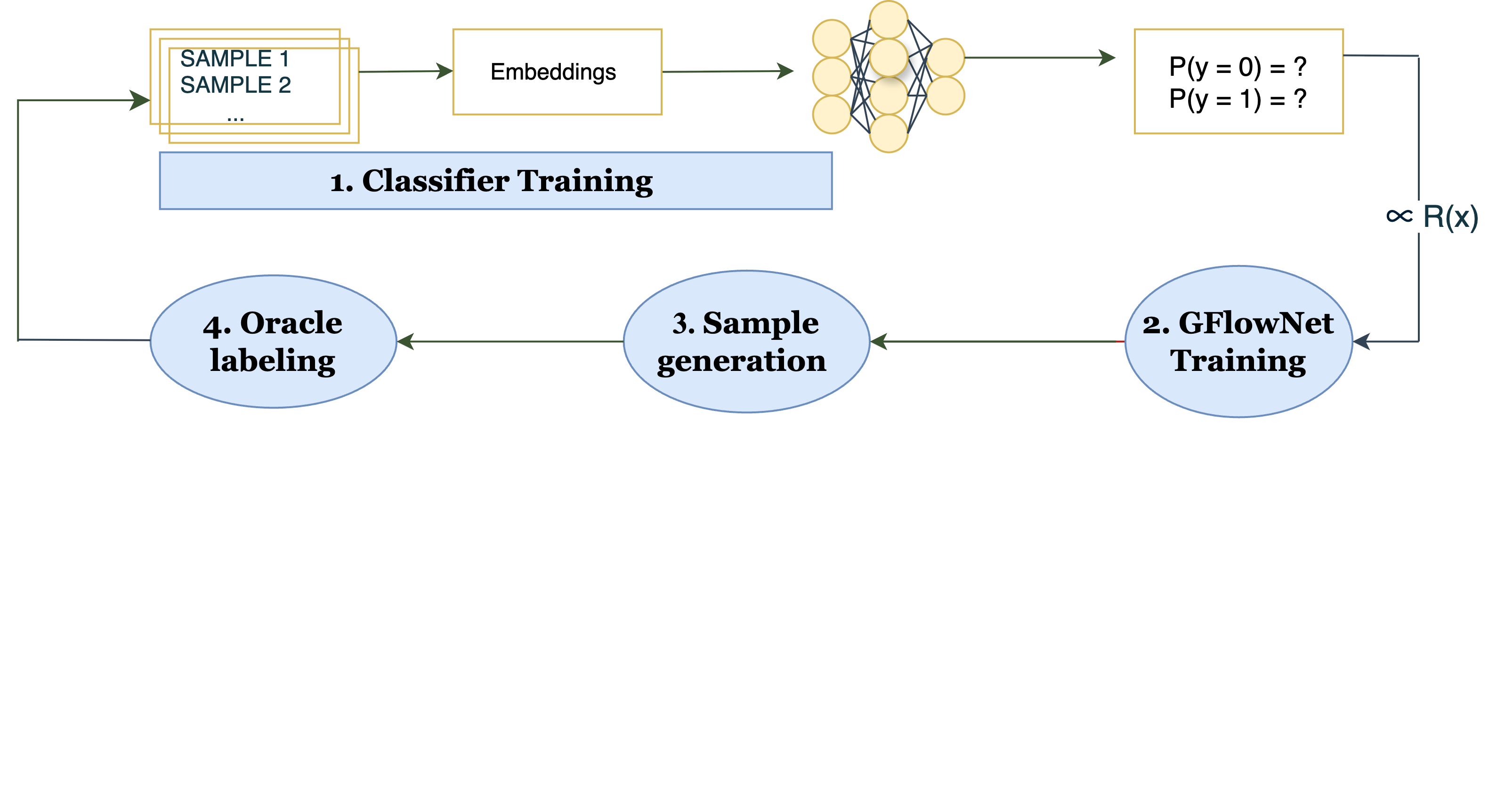} 
  \caption{The generative active learning pipeline for BALD-GFlowNet begins by training a binary classifier on an initial training dataset. This classifier is subsequently used in GFlowNet for computing the mutual information to guide the generation of a batch of samples where the classifier’s predictions are most uncertain. These newly generated samples are labeled by an oracle and are subsequently added to the training dataset to retrain the classifier and improve its performance in the next iteration.}
\label{fig:framework}
\end{figure}

\section{Background and Related Work}
\subsection{Active Learning}
Uncertainty-based approaches, such as Bayesian Active Learning by Disagreement (BALD) \cite{houlsby2011bayesian}, select the most informative samples with the highest predictive disagreement among the model ensemble. 
Data points with high mutual information are those where the average prediction is uncertain, but individual models sampled from the posterior are confident. While effective at identifying individual informative samples, such methods do not scale well to large datasets, as they require exhaustively computing the MI score for every sample in the unlabeled pool. 



\subsection{GFlowNets in Drug Discovery}
Recent studies have demonstrated the effectiveness of GFlowNets in molecular design and drug discovery, due to their ability to generate diverse, reward-aligned samples. For instance, \cite{jain2022biological} and \cite{ korablyov2024generative} have applied GFlowNets to de-novo design of biological sequences and discovery of novel protein binders. \cite{pandey2024gflownet} demonstrated that atom-level GFlowNets, with rewards combining drug likeliness metrics 
can generate chemically feasible molecules \cite{ertl2009estimation, prasanna2009topological, tian2015application}. 




\section{BALD-GFlowNet Framework}

\textbf{BALD-GFlowNet} is an active learning framework that integrates the BALD objective with GFlowNets to learn a policy for generating informative samples. Instead of exhaustively computing an acquisition score for all objects in the unlabeled pool, our method learns a generative policy to sample them directly. As shown in Algorithm~\ref{alg:gfn_al_loop} and Figure~\ref{fig:framework}, the process begins by fitting a surrogate model to a small, initial training dataset. This model then computes a reward for each sample using the BALD acquisition score \cite{houlsby2011bayesian}. We use this principled heuristic because it seeks samples on which models from the posterior distribution are most likely to disagree.  A GFlowNet is subsequently trained to sample objects with probability proportional to this BALD reward. The generated candidates are then labeled by an oracle and added to the training set. The surrogate model is updated with this new data, and the cycle repeats. This iterative framework maintains a constant acquisition cost per cycle, making BALD-GFlowNet scalable to billion-sized pools.

\begin{algorithm}[H]
\caption{BALD-GFlowNet Active Learning Framework}
\label{alg:gfn_al_loop}
\begin{algorithmic}[1]   
\Require
    Initial training set size $N_0$; Unlabeled pool of objects $\mathcal{D}_{\text{pool}}$; Surrogate Model $f$
\Require 
    GFlowNet training episodes $T$; AL iterations $N$; Acquisition batch size $k$
\State $\mathcal{D}_{\text{train}} \gets 
\text{SampleInitialData}(\mathcal{D}_{\text{pool}}, N_0)$ 
\State $\mathcal{D}_{\text{pool}} \gets \mathcal{D}_{\text{pool}} \setminus \mathcal{D}_{\text{train}}$
\State Initialize a GFlowNet Policy $\pi$
\State Train an initial surrogate model $f$ with $\mathcal{D}_{\text{train}}$
\Statex
\For{$t = 1, \dots, N$}
    \For{episode = $1, \dots, T$}
    \State $\pi \gets \text{TrainGFNPolicy}(\pi, \text{BALD})$ 
    
    \EndFor
    \Statex
    \State $ b_t \gets \text{SampleFromGFNPolicy}(\pi, k)$ 
    \State $y_t \gets \text{QueryOracle}(b)\ \forall b \in b_t$
    \State $\mathcal{D}_{\text{train}} \gets 
    \mathcal{D}_{\text{train}} \cup \{ (b_t, y_t) \}$
    \State $\mathcal{D}_{\text{pool}} \gets \mathcal{D}_{\text{pool}} \setminus \mathcal{D}_{\text{train}}$
    \State Update the surrogate model $f$ on $ \mathcal{D}_{\text{train}}$ 
\EndFor
\State \Return Trained surrogate model $f$
\end{algorithmic}
\end{algorithm}

\section{Empirical Results}
\label{sec:results}
We first evaluate our proposed framework, BALD-GFlowNet, on a synthetic two-dimensional (2D) grid that is small enough to allow for exact analysis, and compare its sampling efficiency against traditional active learning baselines. We find that BALD-GFlowNet (i) effectively navigates the grid to sample points with high uncertainty, (ii) identifies informative data points more efficiently than an exhaustive BALD search. We apply BALD-GFlowNet to a large-scale virtual screening domain to improve the performance of a classifier that predicts the binding affinity of small molecules to a target protein. Here, the GFlowNet agent directly generates molecules for acquisition, guided by a composite reward function that combines the BALD score with key drug-likeness properties. We find that BALD-GFlowNet samples high-reward molecules faster than baselines, when scaling to large chemical spaces with billions of compounds.

\subsection{2D Grid Experiment}

To rigorously evaluate our proposed method in a controlled setting, we first verify its performance on a synthetic task with 2D grid. This grid environment allows for a precise analysis of the GFlowNet-based policy’s sampling and computational scalability and comparison against established active learning baselines.





The BALD-GFlowNet framework here consists of a surrogate model and an active learning agent learning the GFlowNet policy. The GFlowNet agent starts at the origin (0,0). It then navigates the grid using an action space comprising four directional moves (up, down, left, right) and a terminal "stop" action. The policy is trained to sample terminal states x with probability proportional to a reward R(x) defined by the BALD acquisition score at that state. 

\begin{figure}[h!]
    \centering
    \includegraphics[width=\textwidth]{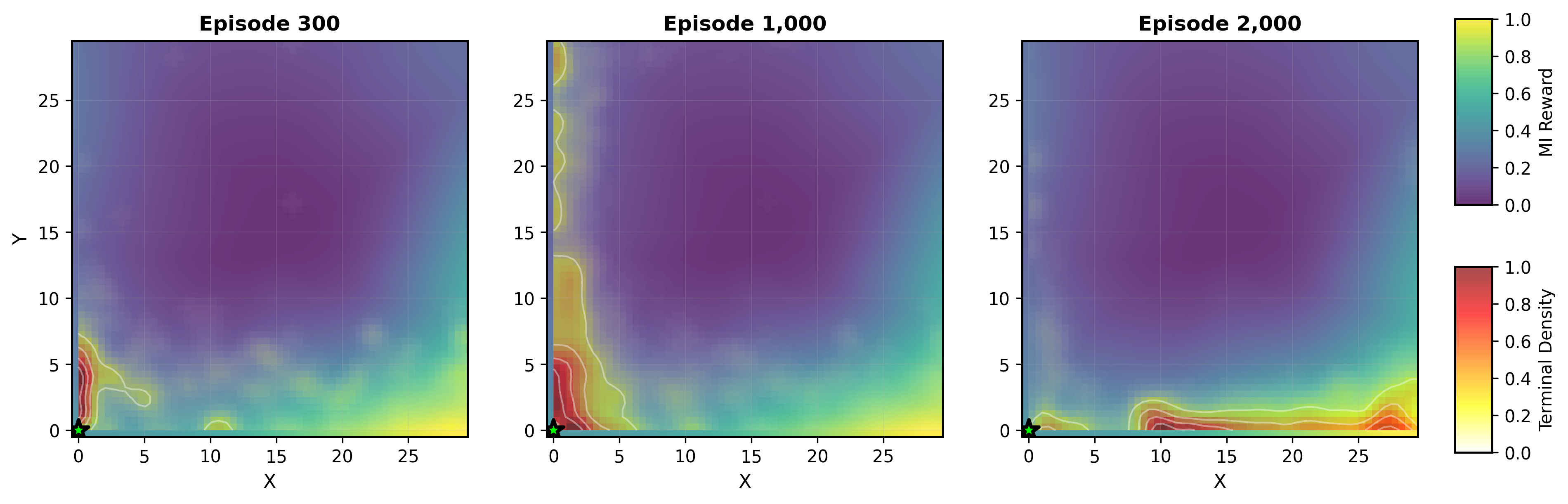}
    \caption{
        \textbf{Terminal State Distribution Over Time}
        The background shows the mutual information (MI) reward landscape (colorbar, top right), with warmer colors indicating higher rewards. Density heatmaps (colorbar, bottom right) represent the distribution of episode terminal states over the most recent 300 trajectories at three training checkpoints: 300, 1000, and 2000 episodes. White contour lines delineate regions of equal terminal density. The green star indicates the initial state position. Early training (left) shows broad exploration, whereas later training stages show convergence toward high-reward regions of the state space.
    }
    \label{fig:terminal_evolution}
\end{figure}

To analyze the GFlowNet's learning behavior and sampling efficiency, we focus on a single active learning step where the reward landscape is fixed. To understand how the GFlowNet's exploration evolves, we analyzed the distribution of terminal states from trajectories sampled at different training episodes. In early training (Figure~\ref{fig:terminal_evolution}, left and middle panels), as the policy learns basic navigation, terminal states are spread across the state space with no clustering patterns. In contrast, later stage (Figure~\ref{fig:terminal_evolution}, right) shows trajectories converging toward regions with high MI reward, indicating more focused and reward-driven exploration.



Figure~\ref{fig:novel_states} shows the efficiency gains of our method. While the BALD baseline requires \(O(n ^2)\) oracle calls with respect to grid size $n$, our method uses substantially fewer calls. Although the number of oracle calls for our method increases with training, it quickly plateaus at a much lower level than the baseline, highlighting its computational efficiency without exhaustive evaluation.

\subsection{Virtual Screening (VS) Case Study}
The objective of our virtual screening case study is to generate a diverse and informative set of molecules to improve docking score classification for Janus Kinase 2 (JAK2) \cite{gabler2013jak2}. To achieve this, we finetune a pretrained GFlowNet \cite{pandey2024gflownet}, which uses a graph-based Transformer policy to construct molecules atom-by-atom, from a chemical action space of atom and bond additions. We then guide the generation process by integrating the BALD objective into the reward function.

To replicate the low hit rate characteristic of VS campaigns, we create a highly imbalanced classification task.  
Then we finetune the GFlowNet using the following composite reward function designed to balance informativeness with drug-likeness properties:

\begin{equation}
\textit{Reward}(x) = \textit{MI}(x) \cdot \textit{TPSA}(x) \cdot \textit{QED}(x) \cdot \textit{SAS}(x) \cdot \textit{Rings}(x).
\end{equation}

Here \(MI(x)\) is the mutual information (\(I(y; \omega \mid x, D)\)) responsible for prioritizing molecules that are informative for the surrogate classifier, while the remaining terms ensure the generated molecules possess desirable chemical properties.

\begin{figure*}[t]
    \centering
    \begin{subfigure}[t]{0.45\textwidth}
        \centering
        \raisebox{2.8pt}{
            \includegraphics[width=\textwidth]{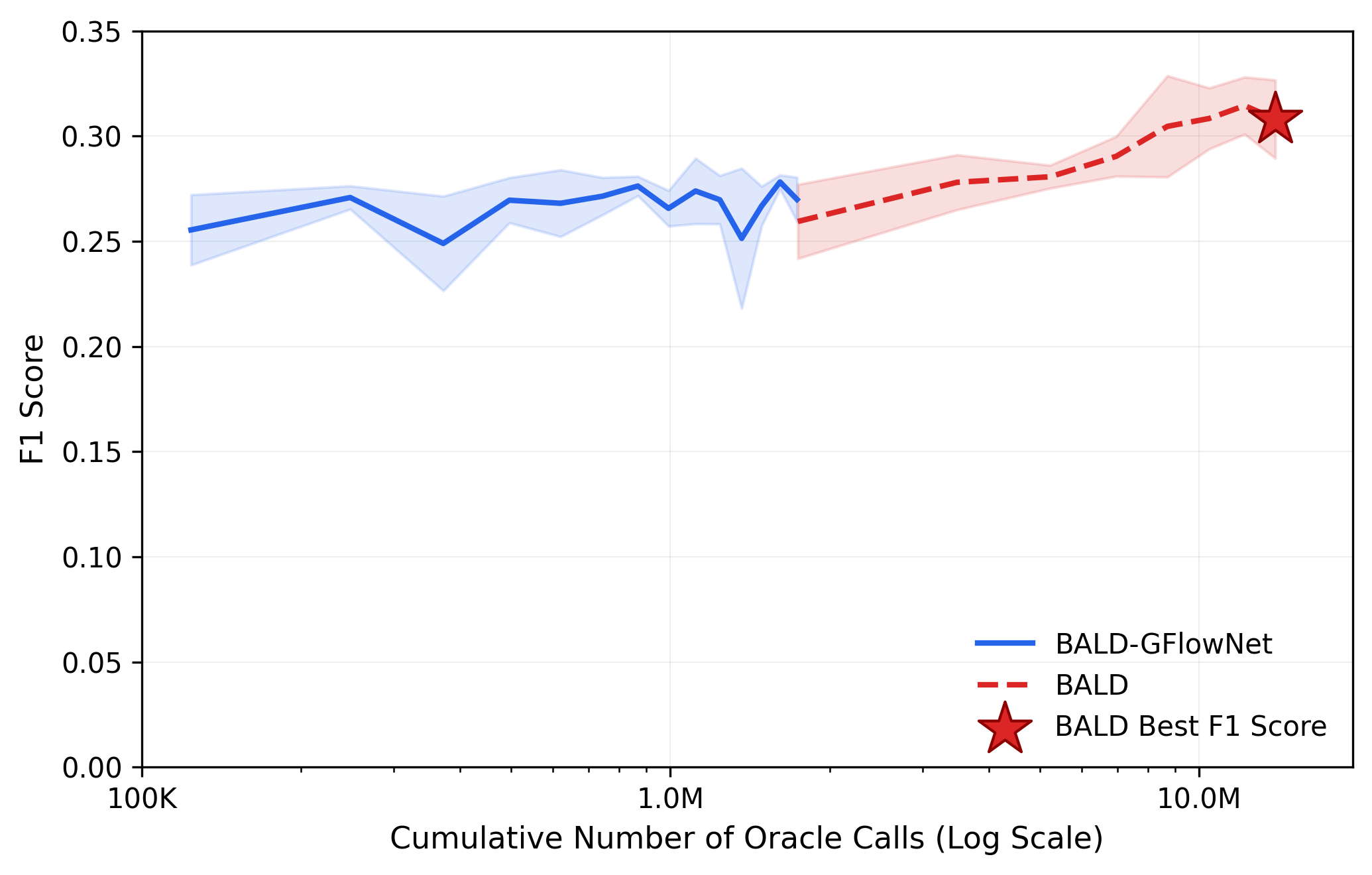}
        }
        \caption{F1 score versus surrogate model evaluations}
        \label{fig:oracle_efficiency}
    \end{subfigure}
    \hfill
    \begin{subfigure}[t]{0.50\textwidth}
        \centering
        \includegraphics[width=\textwidth]{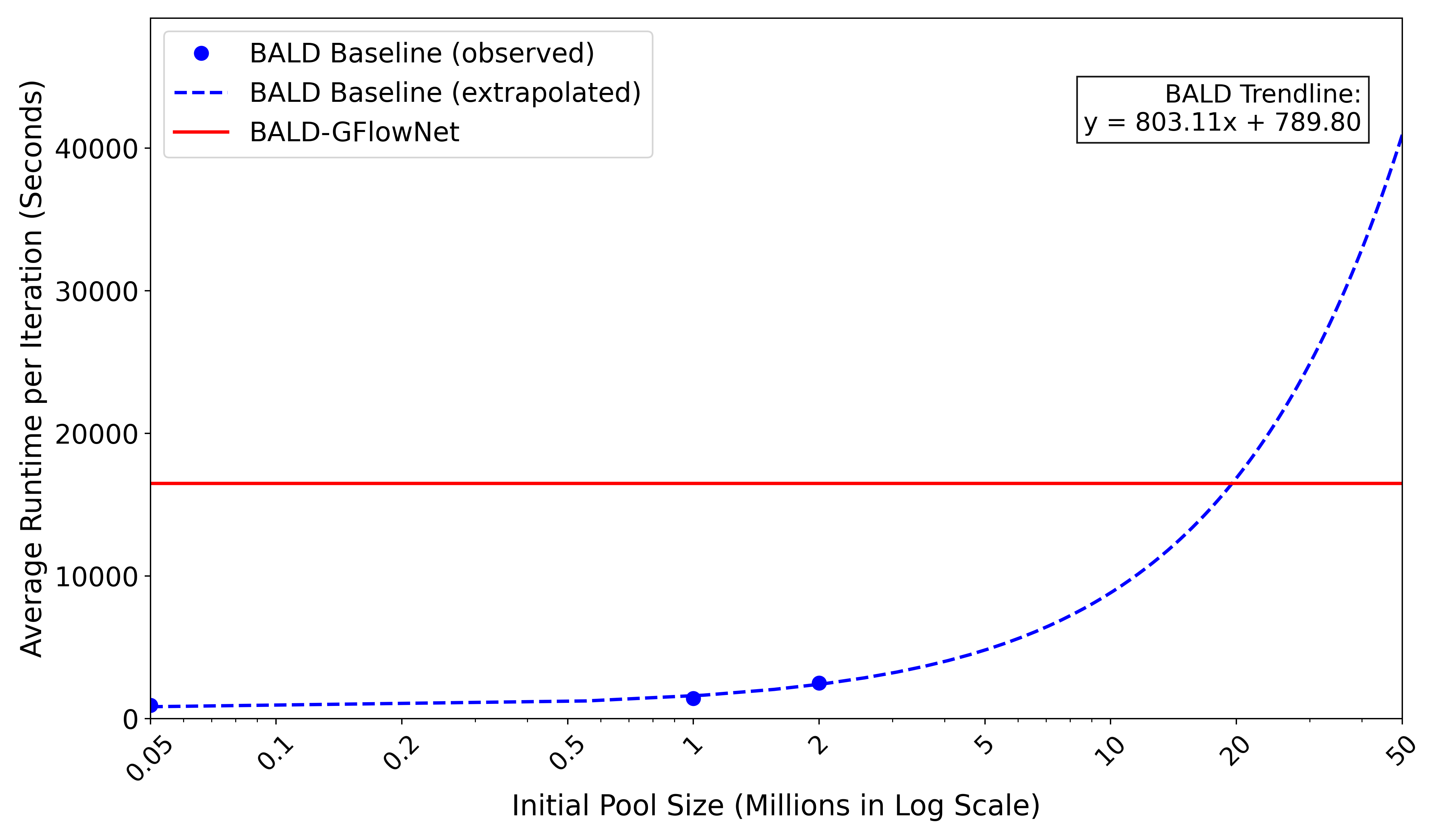}
        \caption{Computational runtime scaling}
        \label{fig:runtime_scaling}
    \end{subfigure}
    \caption{\textbf{Efficiency analysis of BALD-GFlowNet.} 
    (a) F1 score versus the cumulative number of calls to the surrogate model (log scale). Random baseline does not query the surrogate model. BALD-GFlowNet (blue) achieves peak F1 score of 0.30 after around 1.74M evaluations (124K per iteration × 14 iterations). In contrast, BALD (red dashed) reaches its peak F1=0.32 at iteration 7, requiring 12.2M evaluations (1.74M per iteration × 7).  The red star marks BALD's peak performance. Gaussian smoothing (\(\sigma\) = 1.5) is applied.
    (b) Runtime scaling analysis across different pool sizes (log scale). Standard BALD shows $\mathcal{O}(n)$ complexity, with blue circles showing empirically measured runtimes and the dashed line representing extrapolated values ($y = 803.11x + 789.80$). BALD-GFlowNet maintains constant $\mathcal{O}(1)$ complexity (red), showing orders of magnitude improvement for large-scale molecular discovery tasks.}
    \label{fig:efficiency_analysis}
\end{figure*}

We compare BALD-GFlowNet to BALD baseline on the JAK2 virtual screening task. Our results show that BALD-GFlowNet provides a scalable solution for active learning in massive molecular libraries without sacrificing performance. The convergence point in Figure~\ref{fig:efficiency_analysis}a highlights the efficiency gain: BALD-GFlowNet reaches over 90\% of BALD's maximum F1 score using the same number of oracle calls that BALD needs for a single iteration (also see Table~\ref{tab:f1_comparison_total}). Although BALD baseline achieves marginally higher peak performance, this gain comes at a cost that scales linearly with library size \(O(n)\). In contrast, BALD-GFlowNet's runtime remains constant, and an extrapolation to a molecule pool of size 50 million leads to an estimated 2.5
times wall-clock speedup (Figure~\ref{fig:efficiency_analysis}b).

\section{Discussion 
\& Limitations}
\label{sec: Discussion}
Our empirical results on both the synthetic grid and large-scale VS tasks show that BALD-GFlowNet is a  scalable alternative to traditional pool-based active learning. The framework achieves classification performance comparable to the strong BALD baseline while providing distinct advantages in sampling efficiency. By learning a generative policy, BALD-GFlowNet's acquisition cost remains constant regardless of the unlabeled pool's size. The framework's scalability was confirmed by the number of oracle calls on the synthetic dataset and 2.5 times runtime reduction on a pool of 50 million molecules.
The success of this generative approach for active acquisition suggests a promising direction for other scientific discovery problems characterized by vast search spaces where exhaustive evaluation is impractical.

Despite its promising results, our framework has several limitations. First, the effectiveness of BALD-GFlowNet depends on the surrogate model quality. A low-quality surrogate can provide a misleading reward signal, which prevents the GFlowNet from identifying informative regions. Second, the performance of BALD-GFlowNet is closely linked to the design of the reward function. In VS, an imbalance between MI–based rewards and drug-likeness metrics can bias the model to generate either uninformative or chemically infeasible molecules. Finally, our current implementation is tailored to drug discovery. Extending BALD-GFlowNet to other domains would require the development of domain-specific GFlowNet policy networks.

\section{Acknowledgments and Funding Disclosure}
This research was enabled by funds provided by a Canadian Institutes of Health Research (CIHR) doctoral award (FRN: FBD-187593) to MP. AC and ME would like to acknowledge the support by their respective NSERC Discovery grants.

\bibliographystyle{unsrt}
\bibliography{references}


\newpage
\appendix
\onecolumn
\section{Technical Appendices and Supplementary Material}

\subsection{Synthetic Dataset}
\label{app:hypergrid-construction}
\begin{figure}[h]
    \centering
    \includegraphics[width=0.48\textwidth]{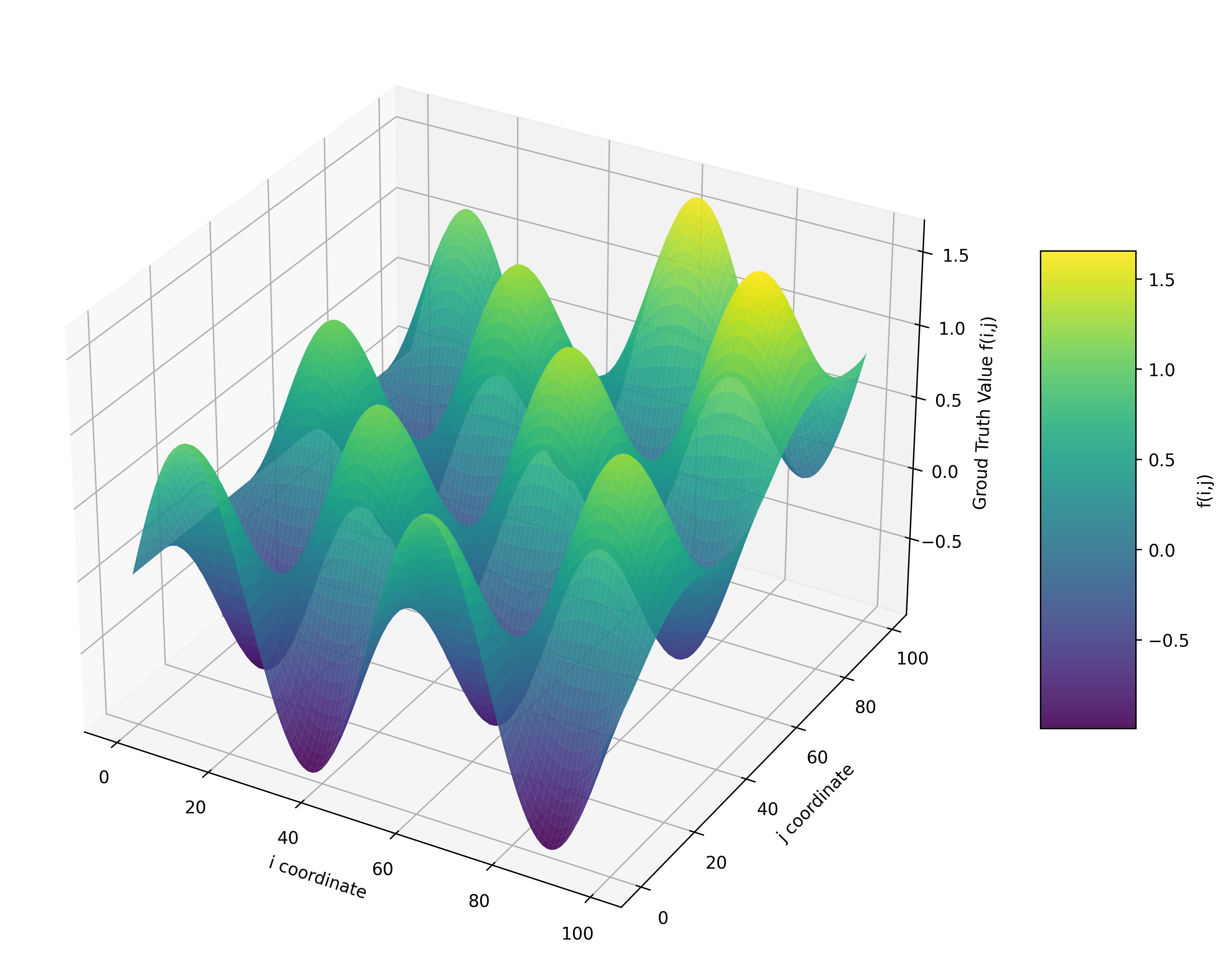}
    \caption{The landscape for ground truth function defined for the 2D grid domain.}
    \label{images/ground_truth_3d_plot.png}
\end{figure}

The 2D 30x30 grid experiment is conducted on dataset constructed by function \( f(i,j) = \sin\!\left(\frac{2\pi i}{50}\right) \cos\!\left(\frac{2\pi j}{50}\right) + \frac{ij}{10000} + \exp\!\left(-\frac{(i-70)^2 + (j-30)^2}{2 \cdot 20^2}\right) + \varepsilon \), where \(\varepsilon \sim \mathcal{N}(0,\,0.1^{2})\) is added as  Gaussian noise (see Figure~\ref{images/ground_truth_3d_plot.png} for the full landscape).
 
\subsection{Synthetic Grid Experimental Details}
The surrogate BNN uses Monte Carlo Dropout for uncertainty estimation. The positive class for the classification task was defined as any point with a ground-truth function value above the 1st percentile of all outputs. The active learning loop was initialized by training the BNN on a set of 100 points sampled uniformly at random. Then it proceeds for 20 iterations, where each iteration involves acquiring 10 new data points.

The GFlowNet policy network is optimized using the Adam optimizer with a Trajectory Balance loss \cite{malkin2022trajectory} (see Table ~\ref{tab:hypergrid} for detail). To provide meaningful and structured spatial embeddings for both the BNN and the GFlowNet policy, we train an autoencoder which takes positional encodings of 2D coordinates as input, using reconstruction and triplet margin loss functions (see ~\ref{app: autoencoder}).

\subsection{Autoencoder Architecture and Training}
\label{app: autoencoder}
\textbf{Positional Encoding and Autoencoder} Instead of directly inputting the two-dimensional coordinates into the autoencoder, they are first transformed a higher-dimensional embedding. We create separate positional encodings for the x and y coordinates using the method from Vaswani et al. \cite{vaswani2017attention}. The two resulting vectors are then concatenated to form the final input for the autoencoder.
As shown Table~\ref {tab:hypergrid}, both the encoder and decoder are MultiLayer Perceptrons consisting of multiple linear layers with ReLU activation \cite{agarap2018deep}. 

\textbf{Loss Function}
The autoencoder is trained with a composite loss function to produce a spatially aware latent space: 
\begin{equation}
\mathcal{L} = \sum_{i=1}^{N} {\| x_i - {\hat x_i} \|^2} + \lambda_{TML} \cdot \sum_{i=1}^{N}{\max\left(0, \| E(x_i) - E({x_i}^p) \|^2 - \| E(x_i) - E({x_i}^n) \|^2 + m\right)}
\end{equation}

The first term is the Mean Squared Error (MSE) between the original coordinate vectors \(x_i\) 
and their reconstructions \(\hat{x}_i\). The second term is a Triplet Margin Loss, which 
encourages the encoder \(E(\cdot)\) to produce a spatially aware latent space \cite{boone2025joint}. This loss 
operates on triplets consisting of an anchor \(x_i\), a positive sample \(x_i^p\), and a 
negative sample \(x_i^n\). See  Table ~\ref {tab:hypergrid} for detail.



\subsection{2D Grid Surrogate Classifier}
\label{sec:appendix4}
The surrogate model is a Bayesian neural network (BNN) modeled as a Multi-Layer Perceptron with Monte Carlo dropout for uncertainty estimation \cite{houlsby2011bayesian, gal2015dropout, charnock2020bayesian}. The network consists of three hidden layers with ReLU activations and Dropout for Bayesian inference \cite{agarap2018deep, srivastava2014dropout}. The final output layer produces logits over the two classes (see Table ~\ref{tab:hypergrid}). The classifier uses F1 accuracy as its performance metric due to the skewed data distribution. 

\subsection{Policy Network Architecture and Training}
\label{sec: appendix5}
The policy network is designed to map a latent state representation to a probability distribution over a discrete action made up of eight directions and a stop action. A linear projection first transforms the latent coordinates into a fixed hidden dimension, followed by a LeakyReLU activation \cite{srivastava2014dropout, xu2015empirical}. The output is then processed by a stack of transformer encoder layers composed of multi-head self-attention and feedforward blocks, regularized with dropout and LeakyReLU for stable training. A final linear layer maps the hidden features to logits over the action space (refer to Table ~\ref{tab:hypergrid}).

\subsection{2D Grid Training Objective and Exploration} The GFlowNet policy is trained using the Trajectory Balance  objective \cite{malkin2022trajectory}. To encourage exploration and prevent policy collapse, we introduce three mechanisms.  First, an \(\epsilon\)-greedy strategy is used during trajectory sampling, allowing the agent to select a random valid action with probability \(\epsilon\). Second, we enforce a minimum trajectory length during training by masking the stop action with probability \(\epsilon_{stop} \). On top of this, we implement depth-aware masking on the stop action to encourage longer trajectories. Stop action is allowed with a probability that increases with the current trajectory's length. In addition, we implement a one-step lookahead that skips any move whose next state would lead to repeated actions or boundaries.

\subsection{Mixup Strategy} To address class imbalance, we apply mixup directly to the learned latent representations \cite{zhang2020does}. In the 2D grid scenario, mixup is performed on encoded spatial embeddings from raw coordinates, whereas in the VS case study, it is applied to molecular embeddings generated by MoLFormer \cite{ross2022large}. Synthetic examples are generated by creating a weighted average of two embeddings within the same class: 
\begin{equation}
\mathbf{v}_{\text{mixup}} = \lambda \cdot \mathbf{v}_1 + (1 - \lambda) \cdot \mathbf{v}_2
\end{equation} where the mixing coefficent \(\lambda \sim {Uniform [0,1]}\). In the case study, the attention mask is formed by the union of the parent masks, preserving the contextual scope of both source molecules.

\subsection{Virtual Screening Implementation Details}
\label{sec: vs appendix}
\textbf{Dataset Acquisition and Processing} Our dataset is derived from the Enamine REAL database, a collection of synthetically feasible molecules well-suited for virtual screening campaigns \cite{shivanyuk2007enamine}. To construct our target-specific dataset, we first selected 2 million of these molecules while using the Deep Docking protocol for Janus Kinase 2 (JAK2) to ensure most potential hits are included in the dataset \cite{gentile2020deep, gabler2013jak2}.

From the dataset, we randomly sampled 10,000 molecules as training dataset and 100,000 molecules as held-out test set, with the remaining compounds forming the unlabeled pool. At each active learning iteration, we actively acquire 100 molecules which are subsequently labeled by QuickVina 2 \cite{alhossary2015fast, lee2023drug}. To adapt the dataset for classification, we convert docking scores into binary labels using a cutoff at the 0.01 quantile of the most negative docking scores in the initial training dataset. The positive class denotes strong binding affinity and the negative class indicates weak or no interaction.   

\textbf{Framework Adaptation for the Virtual Screening Task} We first train a surrogate classifier that adopts a MoLFormer-based neural network, on a dataset of 10,000 randomly sampled molecules from the Enamine REAL database \cite{shivanyuk2007enamine, ross2022large}. Docking scores are converted to binary labels using a stringent 1\% cutoff, and we use mixup technique to handle the resulting imbalance so the model can generalize better to rare small molecules with high affinity 
\cite{zhang2020does}. The classifier is then re-trained in each iteration of the generation process. 


We then compare three acquisition strategies: (i) BALD, which exhaustively scores all unlabeled points and selects the top-k by mutual information \cite{houlsby2011bayesian}; (ii) Random, which selects points uniformly from the unlabeled pool; and (iii) GFlowNet, which samples points from a trained generative policy. 

\textbf{GFlowNet Fine-tuning and Regularization} To further maintain the feasibility of generated compounds, we regularize the finetuning process with a Relative Trajectory Balance (RTB) objective and an offline dataset of feasible molecules \cite{pandey2024gflownet, venkatraman2024amortizing}.

\textbf{Surrogate Classifier} Our surrogate model utilizes the MoLFormer encoder architecture to generate embeddings from molecular SMILES strings \cite{ross2022large}. These tokenized embeddings are then processed by a classification head, consisting of Multi-Layer Perceptron and Multi-Head Attention layers, to yield a final prediction. First, the attention pooling layer assigns a different weight to each token, creating a unifying contextualized representation of the molecule \cite{vaswani2017attention}. Then this pooled representation goes through a DenseNet-like architecture that allows the model to build progressively more complex and hierarchical features \cite{huang2016densely}. The refined output passes through a self-attention mechanism to refine its representation by re-weighing the importance of different internal features. See Table~\ref{tab:hyperparameters} for a summary of the model hyperparameters.

\section{Additional Plots and Results}


\subsection{Oracle Call Comparison on the Synthetic Grid Task}
\begin{figure}[h]
\centering
\includegraphics[scale=0.4]{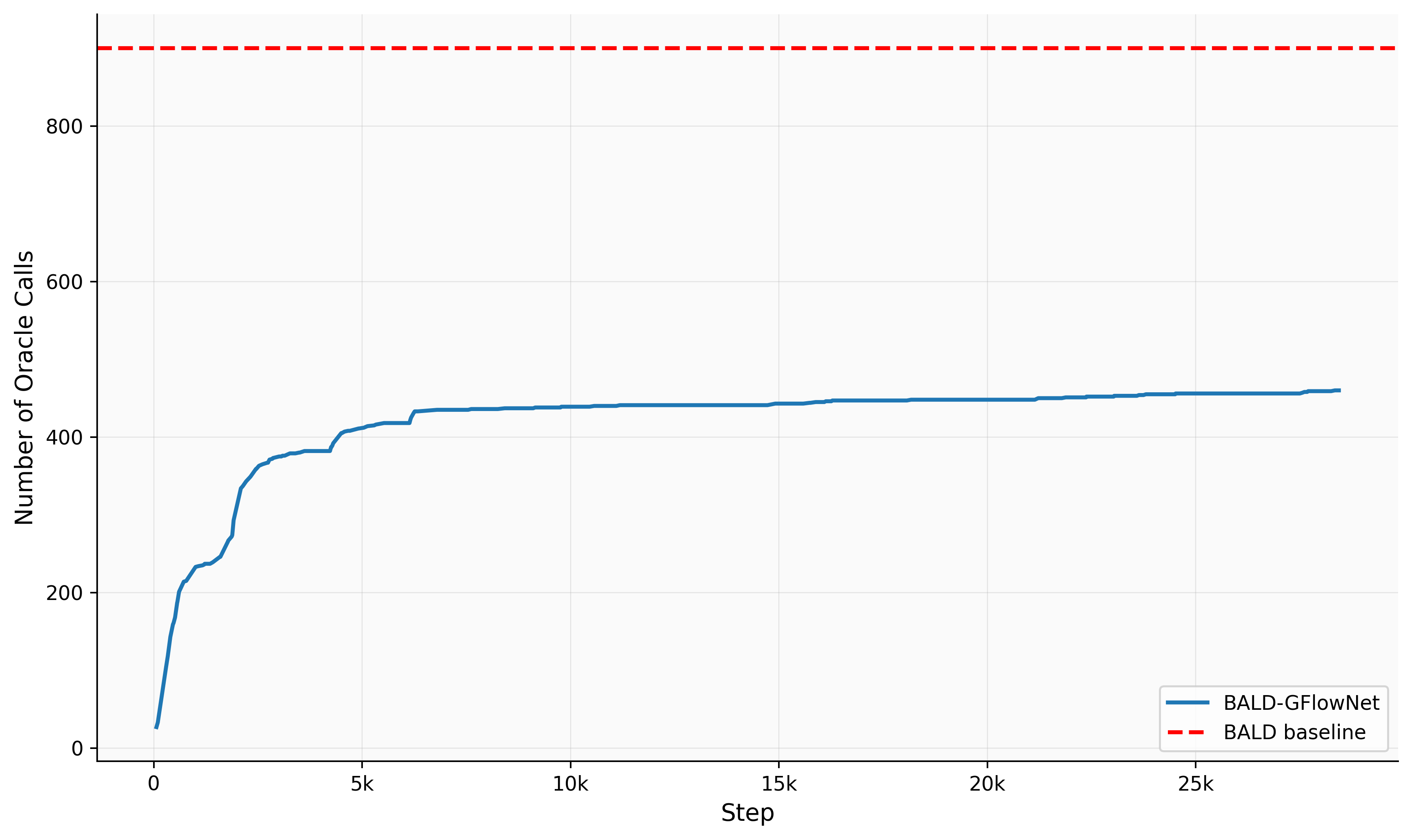}
\caption{Oracle calls for BALD-GFlowNet and the BALD baseline in the 2D grid environment for a single AL step. The Random baseline does not query the surrogate to compute acquisition rewards for selected samples, so its call count is zero. }
\label{fig:novel_states}
\end{figure}

Figure~\ref{fig:novel_states} compares the number of oracle calls between BALD-GFlowNet and the BALD baseline. Here the number of oracle calls is defined as the cumulative count of unique terminal states selected for labeling, with each state corresponding to a unique query to the surrogate model. Compared to the baseline's constant \(O(n)\) calls, BALD-GFlowNet requires much fewer oracle calls.

\subsection{Peak F1 Score and Query Number Comparison on VS Task}
This table summarizes the peak F1 Score against the total computational cost, measured in cumulative number of queries to the oracle required to reach that peak. BALD-GFlowNet achieves a competitive F1 score while being approximately 7 times more computationally efficient than the standard BALD baseline.


\begin{table}[h]
\centering
\caption{Peak performance and acquisition queries for each strategy.}
\label{tab:f1_comparison_total}
\begin{tabular}{lccc}
\toprule
\textbf{Method} & \textbf{Peak F1} & \textbf{Peak Iteration} & \textbf{Queries at Peak} \\
\midrule
\textbf{BALD-GFlowNet} & 0.30 & 14 & 1.74M \\
BALD & 0.32 & 7 & 12.2M \\
Random & 0.28 & 5 & 0 \\
\bottomrule
\end{tabular}
\end{table}



\clearpage
\begin{table}[H]
\centering
\caption{Hyperparameter configuration for models in the grid experiment.}
\label{tab:hypergrid}
\begin{tabular}{ll}
\toprule
\textbf{Hyperparameter} & \textbf{Value} \\
\midrule
\multicolumn{2}{c}{\textit{Active Learning Pipeline}} \\
\midrule
Total AL Steps & 20 \\
Initial Dataset Size & 100 \\
Acquisition Size per Step & 10 \\
Test Set Size & 100 \\
Grid Size & 30 $\times$ 30 \\
\midrule
\multicolumn{2}{c}{\textit{Coordinate Autoencoder}} \\
\midrule
Positional Encoding Dimension & 128 \\
Number of Layers (Encoder/Decoder) & 6 \\
Hidden Dimension & 512 \\
Latent Dimension & 50 \\
Activation Function & ReLU \cite{agarap2018deep} \\
Loss Function & MSE + Triplet Margin Loss \cite{boone2025joint} \\
Triplet Loss Weight ($\lambda_{\text{TML}}$) & 0.1 \\
Triplet Loss Margin ($m$) & 1.0 \\
Optimizer & Adam \cite{kingma2014adam} \\
Learning Rate & 1e-3 \\
Training Epochs & 50 \\
Batch Size & 512 \\
\midrule
\multicolumn{2}{c}{\textit{Surrogate BNN Model}} \\
\midrule
Input Dimension & 50  \\
Hidden Dimension & 256 \\
Number of Hidden Layers & 1 \\
Dropout Rate & 0.1 \\
Number of Dropout Models & 3\\
Optimizer & Adam \cite{kingma2014adam} \\
Learning Rate & 1e-3 \\
Weight Decay & 1e-4 \\
Training Epochs & 7 \\
Loss Function & Cross Entropy Loss \cite{zhang2018generalized}\\
\midrule
\multicolumn{2}{c}{\textit{GFlowNet Policy (Transformer)}} \\
\midrule
Input Dimension & 50 \\
Hidden Dimension & 256 \\
Number of Encoder Layers & 6 \\
Number of Attention Heads & 8 \\
Feed-Forward Dimension & 1024 \\
Dropout Rate & 0.1 \\
Number of Actions & 5 \\
Optimizer & Adam \cite{kingma2014adam}\\
Learning Rate & 1e-4 \\
Loss Function & Trajectory Balance (TB) \cite{malkin2022trajectory}\\
Training Episodes per AL Step & 50000 \\
Min Trajectory Length & 50 \\
Max Trajectory Length & 100 \\
$\epsilon$-Greedy & 0.1 \\
$\epsilon$-Stop & 0.5 \\
Initial Partition & 10 \\
\bottomrule
\end{tabular}
\end{table}


\clearpage
\begin{table}[h]
\centering
\caption{Hyperparameter configuration for models involved in the BALD-GFlowNet pipeline.}
\label{tab:hyperparameters}
\begin{tabular}{ll}
\toprule
\textbf{Hyperparameter} & \textbf{Value} \\
\midrule
\multicolumn{2}{c}{\textit{BALD-GFlowNet Pipeline}} \\
\midrule
Active Learning Iterations & 30 \\
Initial Dataset Size & 10000 \\
Acquisition Size & 100 \\
Initial Pool Size & 1741777\\
\midrule
\multicolumn{2}{c}{\textit{GFlowNet Model}} \\
\midrule
Loss Function & RTB \cite{venkatraman2024amortizing}\\
Loss Coefficient & 0.04 \\
MLE Coefficient & 20 \\
MI Reward Range & $[0,1]$ \\
MI Task Slope & 1 \cite{pandey2024gflownet}\\
Beta & 96 \\
Number of GNN Layers & 15 \\
Number of Heads per GNN Layer & 4 \\
Number of Hidden Dimensions & 128 \\
Number of MLP Layers & 10 \\
Sampling Temperature $T$ & 1 \\
Optimizer & Adam \citep{kingma2014adam} \\
Learning Rate & 1e-6  \\
Weight Decay & 1e-8 \\
Momentum & 0.9 \\
Second Momentum & 0.999 \\
Epsilon & 1e-8 \\
Batch Size & 64 \\
Training Steps & 3881 \\
\midrule
\multicolumn{2}{c}{\textit{Classifier Model}} \\
\midrule
Loss Function & Cross Entropy Loss \cite{zhang2018generalized} \\
Training Epochs & 100 \\
Batch Size & 1024 \\
Optimizer & Adam \citep{kingma2014adam} \\
Patience & 10 \\
Learning Rate & 0.000001 \\
Learning Rate Scheduler & gamma 0.1, step size 40 \\
Weight Decay & 0.01 \\
Base Model & MoLFormer \cite{ross2022large}\\
Number of MC Dropout Models & 3 \\
Dropout Rate & 0.1 \\
Number of Dense Layers & 3 \\
Hidden Dimensionality & 768 \\
Intermediate Dimensionality & 3072\\
Multi-head Attention Heads & 8 \\
Residual Blocks & 2 \\
\bottomrule
\end{tabular}
\end{table} 

\section{Additional Information}
\subsection{Compute Resources}
\label{sec:compute}
All experiments were conducted on a private institutional GPU cluster managed by the Slurm workload manager using NVIDIA A100-PCIE-40GB GPUs (40,960 MiB memory, CUDA 12.4). The three methods from the virtual screening case study used 3 GPUs for 13 days, requiring approximately 25,800 MiB of GPU memory. The hypergrid experiment ran on a single GPU for 1 day with memory usage of 5,000 MiB. In total, the reported experiments required approximately 40 GPU-days of computation. No significant additional compute was used for unreported preliminary experiments.

\subsection{License}
The table below provides a list of softwares, datasets, packages and similar with their license and terms of use:

\begin{table}[h]
\centering
\caption{Licenses and terms of use for assets used in this work.}
\label{tab:licenses}
\begin{tabular}{lll}
\toprule
\textbf{Asset} & \textbf{License} & \textbf{Terms of Use / License URL} \\
\midrule
IBM MoLFormer & Apache-2.0 & \url{https://www.apache.org/licenses/LICENSE-2.0} \\
Enamine REAL DB & Not Applicable & \url{https://enamine.net/terms-of-use} \\
QuickVina2 & Apache-2.0 & Not Applicable \\
PyTorch & BSD Style License & \url{https://github.com/pytorch/pytorch/blob/main/LICENSE} \\
RDKit & BSD 3-Clause  & \url{https://github.com/rdkit/rdkit/blob/master/license.txt} \\
\bottomrule
\end{tabular}
\end{table}


\newpage
\section*{NeurIPS Paper Checklist}

\begin{enumerate}

\item {\bf Claims}
    \item[] Question: Do the main claims made in the abstract and introduction accurately reflect the paper's contributions and scope?
    \item[] Answer: \answerYes{}
    \item[] Justification: We claim that our method BALD-GFlowNet achieves comparable F1 score with significantly fewer oracle calls than the BALD baseline. We confirm this claim with experiments on the two-dimensional grid and virtual screening case study (see section~\ref{sec:results}).
    \item[] Guidelines:
    \begin{itemize}
        \item The answer NA means that the abstract and introduction do not include the claims made in the paper.
        \item The abstract and/or introduction should clearly state the claims made, including the contributions made in the paper and important assumptions and limitations. A No or NA answer to this question will not be perceived well by the reviewers. 
        \item The claims made should match theoretical and experimental results, and reflect how much the results can be expected to generalize to other settings. 
        \item It is fine to include aspirational goals as motivation as long as it is clear that these goals are not attained by the paper. 
    \end{itemize}

\item {\bf Limitations}
    \item[] Question: Does the paper discuss the limitations of the work performed by the authors?
    \item[] Answer: \answerYes{} 
    \item[] Justification: Our paper discusses that our method depends on the quality of the surrogate model. Additionally, under the virtual screening case study, our method requires a balance between the BALD reward and drug likeliness metrics (see section~\ref{sec: Discussion}).
    \item[] Guidelines:
    \begin{itemize}
        \item The answer NA means that the paper has no limitation while the answer No means that the paper has limitations, but those are not discussed in the paper. 
        \item The authors are encouraged to create a separate "Limitations" section in their paper.
        \item The paper should point out any strong assumptions and how robust the results are to violations of these assumptions (e.g., independence assumptions, noiseless settings, model well-specification, asymptotic approximations only holding locally). The authors should reflect on how these assumptions might be violated in practice and what the implications would be.
        \item The authors should reflect on the scope of the claims made, e.g., if the approach was only tested on a few datasets or with a few runs. In general, empirical results often depend on implicit assumptions, which should be articulated.
        \item The authors should reflect on the factors that influence the performance of the approach. For example, a facial recognition algorithm may perform poorly when image resolution is low or images are taken in low lighting. Or a speech-to-text system might not be used reliably to provide closed captions for online lectures because it fails to handle technical jargon.
        \item The authors should discuss the computational efficiency of the proposed algorithms and how they scale with dataset size.
        \item If applicable, the authors should discuss possible limitations of their approach to address problems of privacy and fairness.
        \item While the authors might fear that complete honesty about limitations might be used by reviewers as grounds for rejection, a worse outcome might be that reviewers discover limitations that aren't acknowledged in the paper. The authors should use their best judgment and recognize that individual actions in favor of transparency play an important role in developing norms that preserve the integrity of the community. Reviewers will be specifically instructed to not penalize honesty concerning limitations.
    \end{itemize}

\item {\bf Theory assumptions and proofs}
    \item[] Question: For each theoretical result, does the paper provide the full set of assumptions and a complete (and correct) proof?
    \item[] Answer: \answerNA{} 
    \item[] Justification: Not applicable, our paper does not involve any theoretical results.
    \item[] Guidelines:
    \begin{itemize}
        \item The answer NA means that the paper does not include theoretical results. 
        \item All the theorems, formulas, and proofs in the paper should be numbered and cross-referenced.
        \item All assumptions should be clearly stated or referenced in the statement of any theorems.
        \item The proofs can either appear in the main paper or the supplemental material, but if they appear in the supplemental material, the authors are encouraged to provide a short proof sketch to provide intuition. 
        \item Inversely, any informal proof provided in the core of the paper should be complemented by formal proofs provided in appendix or supplemental material.
        \item Theorems and Lemmas that the proof relies upon should be properly referenced. 
    \end{itemize}

    \item {\bf Experimental result reproducibility}
    \item[] Question: Does the paper fully disclose all the information needed to reproduce the main experimental results of the paper to the extent that it affects the main claims and/or conclusions of the paper (regardless of whether the code and data are provided or not)?
    \item[] Answer: \answerYes{}
    \item[] Justification: The synthetic dataset from the grid experiment and the dataset used for virtual screening are easy to generate and we describe how to generate it in the Appendix~\ref{app:hypergrid-construction} and ~\ref{sec: vs appendix}. For the surrogate model architecture and policy networks in both cases, we describe them in detail (Appendix~\ref{sec:appendix4}, ~\ref{sec: appendix5}, and ~\ref{sec: vs appendix}) with hyperparameters from Table~\ref{tab:hypergrid} and Table~\ref{tab:hyperparameters}. We are planning to release the datasets and Github repository shortly.
    
    \item[] Guidelines:
    \begin{itemize}
        \item The answer NA means that the paper does not include experiments.
        \item If the paper includes experiments, a No answer to this question will not be perceived well by the reviewers: Making the paper reproducible is important, regardless of whether the code and data are provided or not.
        \item If the contribution is a dataset and/or model, the authors should describe the steps taken to make their results reproducible or verifiable. 
        \item Depending on the contribution, reproducibility can be accomplished in various ways. For example, if the contribution is a novel architecture, describing the architecture fully might suffice, or if the contribution is a specific model and empirical evaluation, it may be necessary to either make it possible for others to replicate the model with the same dataset, or provide access to the model. In general. releasing code and data is often one good way to accomplish this, but reproducibility can also be provided via detailed instructions for how to replicate the results, access to a hosted model (e.g., in the case of a large language model), releasing of a model checkpoint, or other means that are appropriate to the research performed.
        \item While NeurIPS does not require releasing code, the conference does require all submissions to provide some reasonable avenue for reproducibility, which may depend on the nature of the contribution. For example
        \begin{enumerate}
            \item If the contribution is primarily a new algorithm, the paper should make it clear how to reproduce that algorithm.
            \item If the contribution is primarily a new model architecture, the paper should describe the architecture clearly and fully.
            \item If the contribution is a new model (e.g., a large language model), then there should either be a way to access this model for reproducing the results or a way to reproduce the model (e.g., with an open-source dataset or instructions for how to construct the dataset).
            \item We recognize that reproducibility may be tricky in some cases, in which case authors are welcome to describe the particular way they provide for reproducibility. In the case of closed-source models, it may be that access to the model is limited in some way (e.g., to registered users), but it should be possible for other researchers to have some path to reproducing or verifying the results.
        \end{enumerate}
    \end{itemize}

\item {\bf Open access to data and code}
    \item[] Question: Does the paper provide open access to the data and code, with sufficient instructions to faithfully reproduce the main experimental results, as described in supplemental material?
    \item[] Answer: \answerYes{} 
    \item[] Justification:We are planning to release the datasets and Github repository shortly.
    \item[] Guidelines:
    \begin{itemize}
        \item The answer NA means that paper does not include experiments requiring code.
        \item Please see the NeurIPS code and data submission guidelines (\url{https://nips.cc/public/guides/CodeSubmissionPolicy}) for more details.
        \item While we encourage the release of code and data, we understand that this might not be possible, so “No” is an acceptable answer. Papers cannot be rejected simply for not including code, unless this is central to the contribution (e.g., for a new open-source benchmark).
        \item The instructions should contain the exact command and environment needed to run to reproduce the results. See the NeurIPS code and data submission guidelines (\url{https://nips.cc/public/guides/CodeSubmissionPolicy}) for more details.
        \item The authors should provide instructions on data access and preparation, including how to access the raw data, preprocessed data, intermediate data, and generated data, etc.
        \item The authors should provide scripts to reproduce all experimental results for the new proposed method and baselines. If only a subset of experiments are reproducible, they should state which ones are omitted from the script and why.
        \item At submission time, to preserve anonymity, the authors should release anonymized versions (if applicable).
        \item Providing as much information as possible in supplemental material (appended to the paper) is recommended, but including URLs to data and code is permitted.
    \end{itemize}

\item {\bf Experimental setting/details}
    \item[] Question: Does the paper specify all the training and test details (e.g., data splits, hyperparameters, how they were chosen, type of optimizer, etc.) necessary to understand the results?
    \item[] Answer: \answerYes{} 
    \item[] Justification: We have provided details about hyperparameter, optimizer type, dataset construction and splits in the Appendix~\ref{app:hypergrid-construction} and ~\ref{sec: vs appendix}, as well as Table~\ref{tab:hypergrid} and ~\ref{tab:hyperparameters}. 
    \item[] Guidelines:
    \begin{itemize}
        \item The answer NA means that the paper does not include experiments.
        \item The experimental setting should be presented in the core of the paper to a level of detail that is necessary to appreciate the results and make sense of them.
        \item The full details can be provided either with the code, in appendix, or as supplemental material.
    \end{itemize}

\item {\bf Experiment statistical significance}
    \item[] Question: Does the paper report error bars suitably and correctly defined or other appropriate information about the statistical significance of the experiments?
    \item[] Answer: \answerYes{} 
    \item[] Justification: In addition the mean F1 score, we have also reported the standard deviation across the three different seeded runs for the virtual  screening case study along with the method for computing the statistics in Figure~\ref{tab:f1_comparison_total}.
    \item[] Guidelines:
    \begin{itemize}
        \item The answer NA means that the paper does not include experiments.
        \item The authors should answer "Yes" if the results are accompanied by error bars, confidence intervals, or statistical significance tests, at least for the experiments that support the main claims of the paper.
        \item The factors of variability that the error bars are capturing should be clearly stated (for example, train/test split, initialization, random drawing of some parameter, or overall run with given experimental conditions).
        \item The method for calculating the error bars should be explained (closed form formula, call to a library function, bootstrap, etc.)
        \item The assumptions made should be given (e.g., Normally distributed errors).
        \item It should be clear whether the error bar is the standard deviation or the standard error of the mean.
        \item It is OK to report 1-sigma error bars, but one should state it. The authors should preferably report a 2-sigma error bar than state that they have a 96\% CI, if the hypothesis of Normality of errors is not verified.
        \item For asymmetric distributions, the authors should be careful not to show in tables or figures symmetric error bars that would yield results that are out of range (e.g. negative error rates).
        \item If error bars are reported in tables or plots, The authors should explain in the text how they were calculated and reference the corresponding figures or tables in the text.
    \end{itemize}

\item {\bf Experiments compute resources}
    \item[] Question: For each experiment, does the paper provide sufficient information on the computer resources (type of compute workers, memory, time of execution) needed to reproduce the experiments?
    \item[] Answer: \answerYes{} 
    \item[] Justification: We mentioned the GPU, type of cluster, memory, storage, estimated total compute time in GPU days as well as the amount of total compute for each individual experiment in subsection~\ref{sec:compute}.
    \item[] Guidelines:
    \begin{itemize}
        \item The answer NA means that the paper does not include experiments.
        \item The paper should indicate the type of compute workers CPU or GPU, internal cluster, or cloud provider, including relevant memory and storage.
        \item The paper should provide the amount of compute required for each of the individual experimental runs as well as estimate the total compute. 
        \item The paper should disclose whether the full research project required more compute than the experiments reported in the paper (e.g., preliminary or failed experiments that didn't make it into the paper). 
    \end{itemize}
    
\item {\bf Code of ethics}
    \item[] Question: Does the research conducted in the paper conform, in every respect, with the NeurIPS Code of Ethics \url{https://neurips.cc/public/EthicsGuidelines}?
    \item[] Answer: \answerYes{} 
    \item[] Justification: We respect the code of ethics listed under NeurIPS. 
    \item[] Guidelines:
    \begin{itemize}
        \item The answer NA means that the authors have not reviewed the NeurIPS Code of Ethics.
        \item If the authors answer No, they should explain the special circumstances that require a deviation from the Code of Ethics.
        \item The authors should make sure to preserve anonymity (e.g., if there is a special consideration due to laws or regulations in their jurisdiction).
    \end{itemize}

\item {\bf Broader impacts}
    \item[] Question: Does the paper discuss both potential positive societal impacts and negative societal impacts of the work performed?
    \item[] Answer: \answerYes{} 
    \item[] Justification:Paper presents a method for improving the scalability of active learning. As a generic method, it could be used in applications with positive or negative societal impacts.
    \item[] Guidelines:
    \begin{itemize}
        \item The answer NA means that there is no societal impact of the work performed.
        \item If the authors answer NA or No, they should explain why their work has no societal impact or why the paper does not address societal impact.
        \item Examples of negative societal impacts include potential malicious or unintended uses (e.g., disinformation, generating fake profiles, surveillance), fairness considerations (e.g., deployment of technologies that could make decisions that unfairly impact specific groups), privacy considerations, and security considerations.
        \item The conference expects that many papers will be foundational research and not tied to particular applications, let alone deployments. However, if there is a direct path to any negative applications, the authors should point it out. For example, it is legitimate to point out that an improvement in the quality of generative models could be used to generate deepfakes for disinformation. On the other hand, it is not needed to point out that a generic algorithm for optimizing neural networks could enable people to train models that generate Deepfakes faster.
        \item The authors should consider possible harms that could arise when the technology is being used as intended and functioning correctly, harms that could arise when the technology is being used as intended but gives incorrect results, and harms following from (intentional or unintentional) misuse of the technology.
        \item If there are negative societal impacts, the authors could also discuss possible mitigation strategies (e.g., gated release of models, providing defenses in addition to attacks, mechanisms for monitoring misuse, mechanisms to monitor how a system learns from feedback over time, improving the efficiency and accessibility of ML).
    \end{itemize}
    
\item {\bf Safeguards}
    \item[] Question: Does the paper describe safeguards that have been put in place for responsible release of data or models that have a high risk for misuse (e.g., pretrained language models, image generators, or scraped datasets)?
    \item[] Answer: \answerNA{} 
    \item[] Justification:  We do not release pretrained language models, image generators, scraped datasets, or similar.
    \item[] Guidelines:
    \begin{itemize}
        \item The answer NA means that the paper poses no such risks.
        \item Released models that have a high risk for misuse or dual-use should be released with necessary safeguards to allow for controlled use of the model, for example by requiring that users adhere to usage guidelines or restrictions to access the model or implementing safety filters. 
        \item Datasets that have been scraped from the Internet could pose safety risks. The authors should describe how they avoided releasing unsafe images.
        \item We recognize that providing effective safeguards is challenging, and many papers do not require this, but we encourage authors to take this into account and make a best faith effort.
    \end{itemize}

\item {\bf Licenses for existing assets}
    \item[] Question: Are the creators or original owners of assets (e.g., code, data, models), used in the paper, properly credited and are the license and terms of use explicitly mentioned and properly respected?
    \item[] Answer: \answerYes{} 
    \item[] Justification: We explicitly mention the licenses of assets with their terms of use in Table~\ref{tab:licenses}.
    \item[] Guidelines:
    \begin{itemize}
        \item The answer NA means that the paper does not use existing assets.
        \item The authors should cite the original paper that produced the code package or dataset.
        \item The authors should state which version of the asset is used and, if possible, include a URL.
        \item The name of the license (e.g., CC-BY 4.0) should be included for each asset.
        \item For scraped data from a particular source (e.g., website), the copyright and terms of service of that source should be provided.
        \item If assets are released, the license, copyright information, and terms of use in the package should be provided. For popular datasets, \url{paperswithcode.com/datasets} has curated licenses for some datasets. Their licensing guide can help determine the license of a dataset.
        \item For existing datasets that are re-packaged, both the original license and the license of the derived asset (if it has changed) should be provided.
        \item If this information is not available online, the authors are encouraged to reach out to the asset's creators.
    \end{itemize}

\item {\bf New assets}
    \item[] Question: Are new assets introduced in the paper well documented and is the documentation provided alongside the assets?
    \item[] Answer: \answerNA{} 
    \item[] Justification: We do not introduce any new assets. 
    \item[] Guidelines:
    \begin{itemize}
        \item The answer NA means that the paper does not release new assets.
        \item Researchers should communicate the details of the dataset/code/model as part of their submissions via structured templates. This includes details about training, license, limitations, etc. 
        \item The paper should discuss whether and how consent was obtained from people whose asset is used.
        \item At submission time, remember to anonymize your assets (if applicable). You can either create an anonymized URL or include an anonymized zip file.
    \end{itemize}

\item {\bf Crowdsourcing and research with human subjects}
    \item[] Question: For crowdsourcing experiments and research with human subjects, does the paper include the full text of instructions given to participants and screenshots, if applicable, as well as details about compensation (if any)? 
    \item[] Answer: \answerNA{} 
    \item[] Justification: We do not perform crowdsourcing experiments and research with human subjects.
    \item[] Guidelines:
    \begin{itemize}
        \item The answer NA means that the paper does not involve crowdsourcing nor research with human subjects.
        \item Including this information in the supplemental material is fine, but if the main contribution of the paper involves human subjects, then as much detail as possible should be included in the main paper. 
        \item According to the NeurIPS Code of Ethics, workers involved in data collection, curation, or other labor should be paid at least the minimum wage in the country of the data collector. 
    \end{itemize}

\item {\bf Institutional review board (IRB) approvals or equivalent for research with human subjects}
    \item[] Question: Does the paper describe potential risks incurred by study participants, whether such risks were disclosed to the subjects, and whether Institutional Review Board (IRB) approvals (or an equivalent approval/review based on the requirements of your country or institution) were obtained?
    \item[] Answer: \answerNA{} 
    \item[] Justification: We do not perform crowdsourcing or experiments related to human subjects.
    \item[] Guidelines:
    \begin{itemize}
        \item The answer NA means that the paper does not involve crowdsourcing nor research with human subjects.
        \item Depending on the country in which research is conducted, IRB approval (or equivalent) may be required for any human subjects research. If you obtained IRB approval, you should clearly state this in the paper. 
        \item We recognize that the procedures for this may vary significantly between institutions and locations, and we expect authors to adhere to the NeurIPS Code of Ethics and the guidelines for their institution. 
        \item For initial submissions, do not include any information that would break anonymity (if applicable), such as the institution conducting the review.
    \end{itemize}

\item {\bf Declaration of LLM usage}
    \item[] Question: Does the paper describe the usage of LLMs if it is an important, original, or non-standard component of the core methods in this research? Note that if the LLM is used only for writing, editing, or formatting purposes and does not impact the core methodology, scientific rigorousness, or originality of the research, declaration is not required.
    \item[] Answer: \answerNA{} 
    \item[] Justification: The core method in this research does not involve LLMs as any important, original, or non-standard components. 
    \item[] Guidelines:
    \begin{itemize}
        \item The answer NA means that the core method development in this research does not involve LLMs as any important, original, or non-standard components.
        \item Please refer to our LLM policy (\url{https://neurips.cc/Conferences/2025/LLM}) for what should or should not be described.
    \end{itemize}

\end{enumerate}


\end{document}